\documentclass[sigconf]{acmart}

\usepackage{balance}
\usepackage{amsmath}
\usepackage{booktabs}
\usepackage{makecell}

\def\BibTeX{{\rm B\kern-.05em{\sc i\kern-.025em b}\kern-.08emT\kern-.1667em\lower.7ex\hbox{E}\kern-.125emX}}

\copyrightyear{2020}
\acmYear{2020}
\setcopyright{acmcopyright}
\acmConference[AIES '20]{Proceedings of the 2020 AAAI/ACM Conference on AI, Ethics, and Society}{February 7--8, 2020}{New York, NY, USA}
\acmBooktitle{Proceedings of the 2020 AAAI/ACM Conference on AI, Ethics, and Society (AIES '20), February 7--8, 2020, New York, NY, USA}
\acmPrice{15.00}
\acmDOI{10.1145/3375627.3375862}
\acmISBN{978-1-4503-7110-0/20/02}

\begin{document}

\fancyhead{}

\title{Joint Optimization of AI Fairness and Utility: A Human-Centered Approach}

\author{Yunfeng Zhang}
\email{zhangyun@us.ibm.com}
\affiliation{
 \institution{IBM Research AI}
  \streetaddress{1101 Kitchawan Rd}
  \city{Yorktown Heights}
  \state{New York}
  \postcode{10598}
}
\author{Rachel K. E. Bellamy}
\email{rachel@us.ibm.com}
\affiliation{
 \institution{IBM Research AI}
  \streetaddress{1101 Kitchawan Rd}
  \city{Yorktown Heights}
  \state{New York}
  \postcode{10598}
}
\author{Kush R. Varshney}
\email{krvarshn@us.ibm.com}
\affiliation{
 \institution{IBM Research AI}
  \streetaddress{1101 Kitchawan Rd}
  \city{Yorktown Heights}
  \state{New York}
  \postcode{10598}
}

\begin{abstract}
  Today, AI is increasingly being used in many high-stakes decision-making applications in which fairness is an important concern. Already, there are many examples of AI being biased and making questionable and unfair decisions. The AI research community has proposed many methods to measure and mitigate unwanted biases, but few of them involve inputs from human policy makers. We argue that because different fairness criteria sometimes cannot be simultaneously satisfied, and because achieving fairness often requires sacrificing other objectives such as model accuracy, it is key to acquire and adhere to human policy makers' preferences on how to make the tradeoff among these objectives. In this paper, we propose a framework and some exemplar methods for eliciting such preferences and for optimizing an AI model according to these preferences.
\end{abstract}

%
%
%

\maketitle

\section{Introduction}
Artificial intelligence (AI) systems are already working alongside humans as trusted advisors in high-stakes decision-making applications such as mortgage lending, hiring, and prison sentencing. In these domains, fairness is an important concern and there are many examples of AI being biased and making arguably unfair decisions. Decisions in these domains are social constructions, and as such need to incorporate multiple viewpoints. Current AI technology does not readily allow for inclusion of viewpoints from those other than the developer of the AI model. The need for deep technical expertise is a barrier to participation for people outside of technical domains. Research is needed to bridge this gap and allow for voices from multiple domains to influence the creation and adoption of fair AI. This is critical for AI that will advise on decisions that have the potential to discriminate against certain populations, such as racial minorities or people with disabilities. 

Machine learning, the most common form of AI today, is, by its very nature, always a form of statistical discrimination. The discrimination becomes objectionable when it places certain privileged groups or individuals at systematic advantage and certain unprivileged groups at systematic disadvantage. In certain situations, such as employment (hiring and firing), discrimination is not only objectionable, but illegal. AI researchers have produced many algorithms that can reduce bias; however, fairness is not a purely algorithmic concept, but a societal one. Ensuring fairness in AI decision making is not just a matter of creating new algorithms that are "fairer"; we argue that it is an iterative and interactive process that involves multiple stakeholders.

If we are to take opinions from different stakeholders about how to handle fairness issues, we must address the conflicts that often arise within them. For example, consider the fairness of an AI system that determines whether to grant bail to a defendant. Arguably, a defendant's definition of fairness is that he or she should not be falsely denied bail, while a judge's top concern is that he or she is not granting bail to people who are likely to recommit a crime or flee. Meanwhile, society as a whole might be concerned about the system being biased against unprivileged groups. The judge's notion of fairness can sometimes conflict with the defendant's since the judge might have to deny bail requests for people who are unlikely to flee or reoffend if the cost of crime is much higher than the cost of jailing people. The judge's notion of fairness can also conflict with the society's priority if reducing crime means denying more bail requests from unprivileged groups. To reconcile these different fairness criteria, stakeholders must agree on a middle ground and the AI decision making system must optimize decisions based on the constraints set by that agreement.

To this end, we propose a decision framework which 1) formally defines fairness metrics and model utility with regards to the task objective; 2) shows the relationship between different fairness metrics and model utility to the policy makers (PMs) so that they could see how these metrics are connected; 3) uses formal decision analysis techniques to elicit PMs' preferential weights on each metric; 4) finds the optimal model setting that satisfies PMs' preferences. The resultant system thus goes beyond traditional bias mitigation algorithms that only seek to mitigate bias on one particular fairness metric. Instead, we take a human-centered and system point of view to approach the problem.

\section{Related Work and Preliminaries}
\label{sec:related}

\subsection{Fairness Criteria and Bias Mitigation Methods}
The taxonomy proposed by \citet{Barocas} divides various fairness criteria into three categories: independence, separation, and sufficiency. These three fairness definitions, particularly the first two, nicely correspond to the different notions of fairness held by philosophers and the society at large. The independence criterion requires that all groups of people receive equal rate of favorable treatment by the decision system. This roughly corresponds to the notion of equity, where everyone, regardless of their differences and background, should receive support to have equal outcome. The separation criterion requires that the false positive rates and the false negative rates are similar across all groups. This roughly corresponds to the notion of equality, which entails that every group is supported at the same level. Lastly, sufficiency requires that the machine learning classifier assigns a score to each person that accurately captures the impact of all attributes including demographics, even if that means some demographic groups receive, on average, lower scores. With many classifiers, if the model is well trained on a large dataset and has high accuracy, the sufficiency criterion is automatically satisfied because they predict probabilities that are well calibrated to an individual's true probability. Thus, in this paper, we primarily focus on the first two fairness criteria. We also focus on binary classification cases for the sake of simplicity. For multi-class classification, the following equations and metrics need to be extended.

Formally, independence requires that
\begin{equation*}
 P(\hat{Y}=1|G=a) = P(\hat{Y}=1|G=b)
\end{equation*}
where $\hat{Y}$ is the classifier's prediction, and $G$ is a demographic variable taken to be a protected attribute in the application domain. Assuming $\hat{Y}=1$ is the favorable outcome, this equation states that the probability of favorable outcome is the same between any two groups $a$ and $b$. Many metrics relate to this notion of fairness, and the two most used are: statistical parity difference, which is the difference between the two sides of the equation, and disparate impact ratio, which is the ratio of probability of favorable outcome for the unprivileged group to that for the privileged group. Often, perfect independence cannot be achieved, and relaxation on this criteria is allowed. Though the degree of relaxation varies from one domain to another and is context-dependent, a legal baseline was set for the disparate impact ratio---it should not be less than 80\% based on the guidelines from the U.S. Equal Employment Opportunity Commission.

The equations for the separation criterion are:
\begin{align*}
 P(\hat{Y}=1|Y=1, G=a) &= P(\hat{Y}=1|Y=1, G=b)\\
 P(\hat{Y}=1|Y=0, G=a) &= P(\hat{Y}=1|Y=0, G=b)
\end{align*}
where $Y$ is the ground truth label. The first equation indicates that the true positive rates are the same across any two groups, while the second equation indicates that the false positive rates are the same. One metric for measuring deviations from this criteria is called average odds difference, which averages the difference in true positive rates and false positive rates between the privileged group and the unprivileged group. So far, there is no legal precedence set on this criterion, and thus it is up to PMs to decide its acceptable range.

Recently, many methods have been developed to reduce biases (cf. \citealp{Bellamy2018} for an overview of these methods and their open source implementation). We choose to follow the method proposed by \citet{Hardt2016}, which mitigate model bias by setting different classification thresholds for different demographic groups. However, we do not use their optimization framework since it was designed to only optimize the model for one fairness metric (equality of opportunity), whereas our framework attempts to jointly optimize multiple model fairness metrics as well as model utility based on PM-defined priorities and constraints.

In addition, most existing debiasing methods seek to optimize only one fairness metric, but \citet{Barocas}, \citet{Chouldechova2017}, and \citet{Kleinberg2016} have shown that different fairness metrics can conflict with each other, and reducing bias on one metric may actually increase bias on another. The bias mitigation method that we propose here does not seek to achieve perfect fairness on any one criterion. Rather, we try to find a model that strikes a balance between independence, separation, and model utility. In order to do that, we need to elicit the PM's policy preferences.

\subsection{Preference Elicitation}
In our framework, the PM decides the relative priorities of independence, separation, and model utility. Model utility is the net benefit of using a model. For example, in many business domains, this could include reduced cost or increased profit, while in criminal justice scenarios, this may include reduced crimes or population imprisoned. Note that model utility itself could be impacted by multiple factors, and needs to be assessed using historical data and assumptions from the PMs. In this paper, we assume that this cost and benefit assessment has all been completed prior to the stage of prioritizing fairness and utility.

The problem of finding the optimal model based on three criteria is a multi-attribute decision problem, and there is a rich literature on this subject in the decision analysis field. One popular method used by many decision support systems is called direct weighting \cite{Dolan}. It starts by asking a decision maker to first assign scores to the criteria based on their priorities (e.g., in a clinical decision scenario, these criteria may be drug effectiveness, cost, and side effects), and then assign scores to each option (e.g. different drugs) to rank them under each criterion. In either phase, the PM first assigns a score of 1 to the lowest ranked item, and then assign scores to other criteria or options based on how many times more important or better they are. All scores are then normalized by dividing the sum of scores in each category. The scores for the criteria become their weights, and the final comparison between options are made based on the weighted combination of scores over all criteria. This method is easy to execute, but it is lacks test-retest reliability as shown by \citet{BOTTOMLEY2001553}.

Many methods have been proposed to improve the direct weighting method. These methods typically try to improve the consistency of the elicited preferences by making changes to the options and asking the PM to rerank them (e.g., SMART, Simple Multi-Attribute Rating Technique). Some also incorporate other weighting methods such as the rank order centroid method (e.g., SMARTER, SMART Exploting Ranks, cf. \cite{Edwards1994}). The framework we propose here will be based on a method called the Analytic Hierarchy Process (AHP, \cite{Saaty1977}). It is one of the most widely used multi-criteria methods today, particularly in medical applications \cite{Liberatore2008}. The main difference between AHP and direct weighting is that AHP requires the PM to make multiple comparisons between each pair of options or criteria. For each pair, the PM is asked which one is better or more important, and rate the level of importance on a one to nine ratio scale. These ratings are then organized into a matrix, and the maximum eigenvalue of the matrix is used as weightings for the criteria or scores for the options. These eigenvalues in effect combines the result of direct pairwise comparisons and the indirect comparisons that they imply. This process also yields a consistency measure which can be used to check whether the PM provided incompatible pairwise ratio scores. As a result, AHP provides more consistent and reliable weights through one unified process.
\section{An Illustrative Example}
The remainder of the paper will introduce our framework using a hypothetical loan approval scenario, in which a binary classifier is used to decide whether to grant a loan to an applicant. The classifier is trained based on a part of the German Credit dataset published on the UCI Machine Learning Repository \cite{Dua:2019}. This data set has 20 attributes including a person's banking records, credit history, employment status, age, gender, etc. We transformed the dataset's label column such that 1 represents that a person was able to repay his or her loan while 0 represents failure to repay the loan. The entire data set has 1000 instances, and we split them into 64\% training set, 16\% validation set, and 20\% test set. The test set is used to compute independence, separation, and utility measures.

We trained a gradient boosting decision tree model on this dataset using the LightGBM framework \cite{LightGBM}. The model achieves an accuracy of 75\% when the classification threshold is set at 0.5. However, this threshold setting does not lead to the highest utility, which is defined as expected average profit per applicant. The data set description indicates that a false positive error---lending money to someone who cannot repay the loan---costs \textbf{five times} as much as the profit earned from a loan fully repaid. Thus, the threshold must be set much higher than 0.5 to reject more people who may not repay the loan.


The main fairness issue of the German Credit dataset is that it leads to classifiers that may disproportionately penalize young people under 26 years old (Age $<= 25$,  cf. \cite{4909197}). Among the 1000 instances in the dataset, 190 were under age 26, and they were 21\% less likely to get the Good credit rating than the rest of the population. In the next section, we will examine how PMs may use our framework to visualize such biases and find potential solutions.

\section{The Fairness-Utility Space}
The first step of our framework is to show the relationships between fairness and utility to the PMs so that they can see how these measures are connected. Similar to \citet{2019arXiv190405419C}, we find visualizations to be a very effective medium to convey such relationships. However, the visualizations we propose here focus more on the end fairness and utility metrics rather than the intermediary statistics. For the loan example scenario, we use statistical parity difference (\textbf{SPD}) to measure independence, and a variant of the average odds difference to measure separation.

We choose SPD instead of disparate impact ratio for measuring independence because ratios are less stable when subjected to numeric optimization since they quickly grow to large values when the denominator becomes small. However, we still use disparate impact ratio to filter out models that do not satisfy the minimum 80\% legal requirement. That is, we remove models that have a disparate impact ratio of less than 0.8 or greater than 1.25. The 1.25 ensures that the privileged group would not be 80\% less likely to receive favorable outcomes than the unprivileged group, which could happen if bias is over-corrected.

Our variant of the average odds difference metric takes into account the weights assigned to false positive and negative error. That is, we compute a weighted average odds difference (\textbf{WAOD}) as follows:
\begin{equation*}
\frac{W_{FP}(FPR_{p} - FPR_{unp}) + W_{TP}(TPR_{unp} - TPR_{p})}{W_{FP} +  W_{TP}}    
\end{equation*}
where $p$ indicates privileged group, $unp$ indicates unpriviledged group, $FPR$ indicates false positive rate, $TPR$ indicates true positive rate, $W_{FP}$ and $W_{TP}$ are weights for these error. When this measure is below zero, it signals bias against the unprivileged group, and otherwise, bias against privileged group. In the loan approval scenario, based on the prior discussion, we set the unprivileged group to be people under 26, $W_{FP}$ to 5, and $W_{TP}$ to 1.

The utility function is domain specific, and thus we do not propose a general definition here. For the loan approval scenario, we hypothesize that the PM measures utility in terms of profit generated per loan, which can be quantified as follows:
\begin{align*}
&E[Profit] (TPR_{p} N_{p} + TPR_{unp} N_{unp})\\
&-E[Cost] (FPR_{p} N_{p} + FPR_{unp} N_{unp} )
\end{align*}
where $E[*]$ represents the expected average profit for a loan fully repaid or the expected average loss for a defaulted loan, and $N$ indicates the number of applicants in a group. Profit and cost are assumed to be elicited from the PM, who can use existing data to calculate them. For this example, we assume the PM set E[Profit] to \$2,000 and E[Cost] to \$10,000 to maintain the 1 to 5 ratio indicated in the data description. This of course overly simplifies the real profit-cost estimation process which likely depends on many factors. In real world applications, one needs to substitute this equation with an actual utility estimation equation.

With the three metrics defined, we randomly sample a range of threshold settings for the classifier, and for each threshold setting, we calculate the independence-separation-utility triplet. Each threshold setting is a pair of thresholds designated for unprivileged and privileged group separately. For example, if we have a threshold setting of 0.6 for young people under 26 year old, and 0.7 for aged people, then young people will be granted a loan only when the classifier's predicted loan-repayment probability is above 0.6, while aged people will be granted a loan when that probability is above 0.7. For the loan example, we randomly sampled 10,000 pairs of thresholds, and then excluded about half of them that had disparate impact ratio outside the [0.8, 1.25] range as discussed earlier.

\begin{figure}
    \centering
    \includegraphics{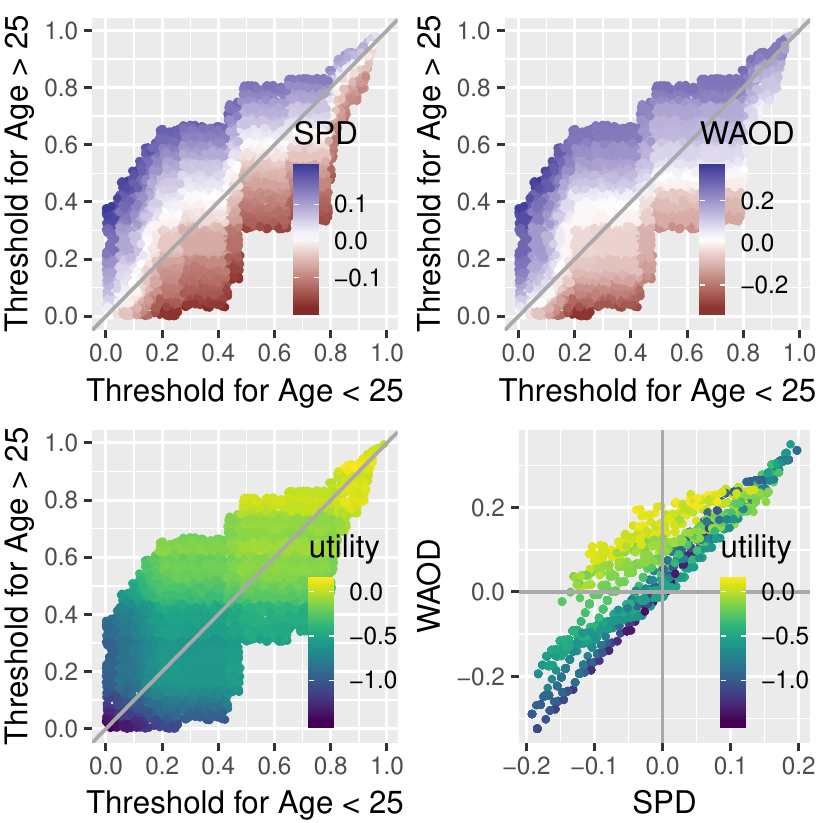}
    \caption{Visualizations showing the relationships among classification threshold settings and the three metrics: statistical parity difference (SPD), weighted average odds difference (WAOD), and utility (in thousands of dollars).}
    \label{fig:german}
\end{figure}

Figure~\ref{fig:german} shows four visualizations we created from the results of the different classification threshold configurations. The top two and the bottom left graphs show directly how changing the thresholds impact SPD, WAOD, and utility. It can be inferred from the top two graphs that the classifier is indeed likely to be biased against young people if the classification thresholds are set to be the same for the young and aged group. This is because the diagonal lines in the two graphs, which are made of equal threshold settings, cross a lot of red regions, which signify negative SPD and WAOD and indicate bias against the unprivileged group. The white regions in these two graphs are where SPD and WAOD are close to 0, indicating fairness according to the respective metric. The SPD graph shows that these white regions are mostly above the diagonal line, meaning that to achieve a fair classifier, in most cases the threshold for aged people has to be set larger than the threshold for young people.

The bottom left graph shows where the highest utility is achieved. As can be seen, utility is mostly negative for many threshold settings, but only become positive when both thresholds are above 0.8 (yellow region at the top right corner). PMs can view graphs like this to see how their cost-benefit assumptions affect the optimal decision threshold from the utility point of view. Combined with the top two graphs, one can also see that this high utility region is close to some white SPD and blue WAOD regions, suggesting that it is possible to find models with high utility and SPD fairness, but that might come with some drawback on the WAOD metric.

The bottom right graph in Figure~\ref{fig:german} removes the thresholds and directly show the relationships among the two fairness and the utility metrics. Ideally, one hopes to find models with the highest utility around the origin point (SPD=0, WAOD=0), but for our example scenario, those points appear to have low utility (dark blue color). In contrast, high utility points (yellow) appeared around WAOD=0.1, though it is unclear from this graph which exact point has the highest utility and close to 0 SPD and WAOD.

\begin{figure}
    \centering
    \includegraphics{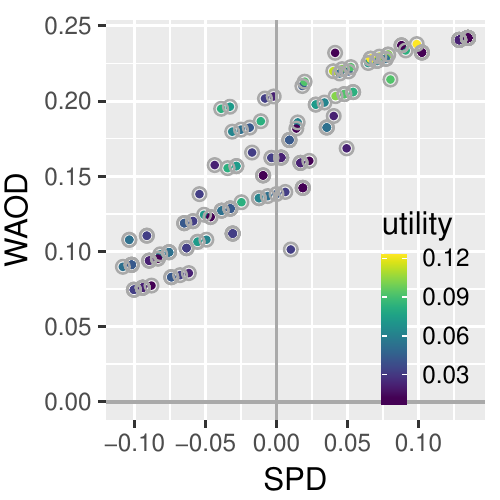}
    \caption{The fairness-utility graph showing only models with positive utility. Utility in thousands of dollars.}
    \label{fig:german_filtered}
\end{figure}

In an application designed to help PMs make decision policies, these graphs can be made interactive to allow PMs filter data based on some metrics. For example, a PM may choose to remove configurations that have negative utility in order to see graphs like the bottom right one in Figure~\ref{fig:german} more clearly. Figure~\ref{fig:german_filtered} shows such a graph, and one can see more clearly where the high utility points lie (e.g. the yellow point around WAOD=0.25 and SPD=0.1). We believe that visualizations like this are important tools to inform PMs about the tradeoff relationships among the three metrics, and is key for them to make the next decision: express their preferences on how to prioritize utility, independence, and separation.

\section{Fairness-Utility Policy Elicitation}
Though the visualizations help PMs see the relationship among independence, separation, and utility, they only show a sample of the possible model configurations. There are likely better configurations outside the 10,000 random samples. To find them, numerical optimization methods such as Bayesian optimization can be deployed, but they require well defined optimization objectives. Therefore, the second step of our framework is to elicit the PM's relative weightings on the fairness and utility metrics so that the optimization objective can be defined. As discussed in the Related Work section, this step is a multi-criteria decision analysis process, and will be completed using a technique called Analytic Hierarchy Process (AHP).

Unlike traditional usage of AHP that requires the PM to rank both the criteria and the alternative options on each criterion, we only ask the PM to rank the criteria. This is because: a) we have infinite amount of options---infinite possible threshold settings for the privileged and unprivileged groups; and b) we do not need the PM to rank each option on each criterion since they can be ranked directly by our independence, separation, and utility metrics. Therefore, as long as we have the PM's weight for each metric, we can combine the three metrics into one single metric and use automatic optimization to find the best model threshold setting.

Using AHP to elicit weightings requires the PM to compare every pair of criterion and the comparisons need to be about a unit of one metric against a unit of another. These units need to be somewhat comparable and intuitively comprehensible for the PMs. One way to determine the unit is to divide the range of each metric into equal number of intervals. In our example, for the model configurations with positive utilities, independence (SPD) ranged from -0.1 to 0.1, separation (WAOD) ranged from 0.5 to 0.25, utility ranged from \$0 to \$200 per applicant. We divided each range into 2 intervals, resulted in an interval size of 0.1 for SPD and WAOD, and \$100 for utility.

\begin{figure}
    \centering
    \includegraphics[width=\columnwidth]{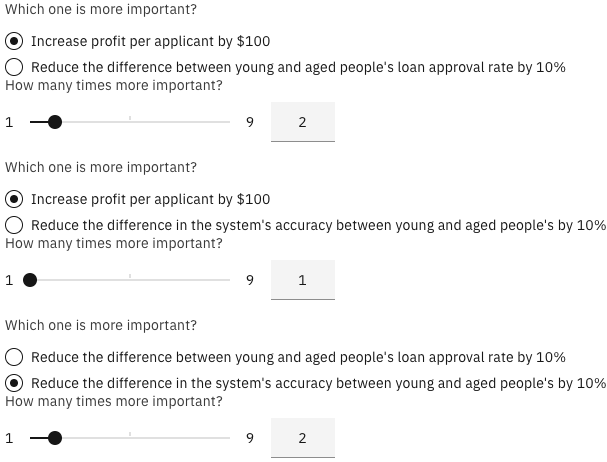}
    \caption{A screenshot of the preference elicitation part of our system.}
    \label{fig:questions}
\end{figure}

Our framework is implemented with an interactive web interface, and Figure~\ref{fig:questions} shows the preference elicitation part that applies the AHP method. Note that in the web interface, we translated the obtuse metric names into domain-based descriptions for PMs to understand. This translation is critical to interface with PMs who may not understand the subtle differences between the two fairness criteria.

On the web interface, the level of importance is assigned on a 1 to 9 scale. For each question, if the PM prioritizes the second option, the importance rating is inverted. The ratings for all three questions are assembled into a matrix:
\begin{displaymath}
\begin{bmatrix} 
1 & a & b\\
1/a & 1 & c\\ 
1/b & 1/c & 1\\
\end{bmatrix}
\end{displaymath}
where $a$, $b$, $c$ represent the ratings assigned to the three questions. AHP then takes the eigenvector for the matrix's largest eigenvalue as the weights for the three metrics.

Using this approach, the PM can express a wide range of preferences. Table~\ref{tab:pref} shows a few example ratings and their resultant weights. The first three cases explore the extreme settings allowed by the AHP ratings. As can be seen, the maximum weight that can be set on any one metric is 0.82 because the maximum rating is limited to 9. The last example showcases the power of AHP expressing some moderate tradeoff preferences. The example says that the PM considers SPD to be as important as utility, and both utility and SPD to be two times as important as WAOD. This results in a weight of 0.4 for utility, 0.4 for SPD, and 0.2 for WAOD, which indeed satisfy the the supplied ratio ratings.
\begin{table*}
  \caption{Some exemplar tradeoff preferences expressed via AHP ratings, their resultant weights for each metric, the thresholds of the optimal model subject to the metric weights, and the final metrics of the optimal model.}
  \label{tab:pref}
  \begin{tabular}{rcccccccccccccc}
    \toprule
    & \multicolumn{3}{c}{AHP ratings} & \multicolumn{3}{c}{Weights} &
    \multicolumn{2}{c}{Optimal Thresholds} & \multicolumn{3}{c}{Metrics}\\
    \cmidrule(lr){2-4}
    \cmidrule(lr){5-7}
    \cmidrule(lr){8-9}
    \cmidrule(lr){10-12}
    & \thead{Util.\\ vs. SPD} & \thead{Util.\\ vs. WAOD} & \thead{SPD\\ vs. WAOD} & $W_{util}$ & $W_{SPD}$ & $W_{WAOD}$ &
    Young & Aged & Util. & SPD & WAOD
    \\
    \midrule
    Max Util. & 9 & 9 & 1 & 0.82 & 0.09 & 0.09 & 95.0\% & 92.7\% & \$130 & -0.06 & 0.11 \\
    Min $|SPD|$ & 1/9 & 1 & 9 & 0.09 & 0.82 & 0.09 & 84.4\% & 85.9\% & \$40 & 0 & 0.14\\
    Min $|WAOD|$ & 1 & 1/9 & 1/9 & 0.09 & 0.09 & 0.82 & 50.9\% & 47.4\% & -\$250 & -0.08 & 0\\
    Balanced & 1 & 2 & 2 & 0.4 & 0.4 & 0.2 & 95.1\% & 94.6\% & \$110 & -0.01 & 0.14\\
  \bottomrule
\end{tabular}
\end{table*}

AHP always run consistency checks on the ratings, thus our system will ask the PM to change their ratings if inconsistency is found. Inconsistency arises when the relationship implied by two comparisons conflicts with the result of the other, direct comparison. For example, if utility is rated to be 9 times as important as SPD, 2 times as important as WAOD, this would imply that WAOD is about 4 times more important than SPD. However, if the PM rates that SPD is more important than WAOD, it would lead to a large inconsistency score. Typically, when the inconsistency score is greater than 0.1, re-rating is needed.

\section{Preference-constrained model optimization}
After the metric weights are elicited, the last step of our framework is to find the optimal model that satisfies the weight constraints. To do this, we first combine the three metrics into an objective function with the elicited weights:
\begin{align*}
-Util\frac{W_{util}}{S_{Util}} + |SPD|\frac{W_{SPD}}{S_{SPD}} + |WAOD|\frac{W_{WAOD}}{S_{WAOD}}
\end{align*}
where $S_{*}$ are the scales applied to each metric in the elicitation questions, i.e. 100, 0.1, and 0.1 for our example. Since this is a minimization objective, we take the negative of utility and the absolute value of SPD and WAOD. In addition, whenever the disparate impact ratio is outside the range [0.8, 1.25], we emit an error for the minimization procedure to avoid violating legal limits.

The goal of the optimization is to find classification threshold settings for the privileged and unprivileged groups that minimize the objective function. However, because the relationship between the three metrics and the threshold settings are non-convex, we cannot use traditional numerical optimization methods. We chose to use Bayesian optimization since it has been successfully used in many hyper-parameter tuning scenarios and have few assumptions about the objective function. Specifically, we use the Tree of Parzen Estimators (TPE) implemented in the python hyperopt library \cite{Bergstra:2013}.

Table~\ref{tab:pref} lists the optimal thresholds found for each preference setting and the resultant scores evaluated on the three metrics. As can be seen, the Bayesian Optimization procedure seems to work very well. The utility for the Max Utility case is indeed the largest of all four options, while Min $|SPD|$ and Min $|WAOD|$ respectively achieved perfect fairness as measured by the two metrics. The result for the last case is interesting because it indeed finds a model that has good utility as well as a close to 0 SPD. This suggests that our framework is effective for PMs who want to find a middle ground between fairness and model utility.





\section{Future Work}
As any method that involves humans in the loop, user studies should be carried out to test the method's effectiveness in practice. This will be the next step of our research. We anticipate some difficulties in helping users understand the two fairness metrics. It is likely that applications based on our framework need to explain the meanings of the fairness metrics to the user repeatedly using domain relevant terms. We also expect that it will be difficult for users to compare the three metrics and assign ratio ratings. We believe that the interactive visualizations we developed is an important step towards helping users make this comparison, but perhaps other forms of visualizations and representations are also needed. Lastly, we will compare AHP with other preference elicitation methods to see which method the users prefer.

Preference elicitation methods like AHP require PMs to make difficult global comparisons. An alternative method that we would like to explore is active-learning styled elicitation, in which the system iteratively asks the PM to make local comparisons between many pairs of options to infer the user's weights on the three metrics. Each comparison question would take the form of comparing one option with $x$ utility, $y$ SPD, and $z$ WAOD to another option with $a$ utility, $b$ SPD, and $c$ WAOD. Based on the PM's answer, the system decides on the next pair of options that would gain maximum information on the user's preference weightings. This method may require asking many questions, but it likely would lead to more reliable preference weightings than the AHP method.

\section{Conclusion}
Fairness is a human construct, and its definition has evolved throughout the entire human history. Because of this, we believe that machine learning systems cannot presume how policy makers would like to handle fairness, but have to elicit fairness criteria from them. To this end, we presented a framework that first elicits people's preferences on how to trade off fairness with model utility, and then efficiently search for the model that maximizes fairness and utility at the same time while respecting the preference constraints imposed by the policy maker. Our framework exhibits an extraordinary capacity in allowing people to express a wide range of preferences and it has succeeded in finding the optimal models.

Although the example we showed here only involved a single policy maker, our framework works for a group of policy makers as well. This is because the Analytic Hierarchy Process (AHP) has been extended to aggregate group preferences \cite{Ossadnik2016}. This makes our framework well suited for real world decision tasks where the stakeholders often have very different backgrounds and interests. For example, for the decisions of granting bail requests, judges may want to prioritize reducing crimes, while citizens may want to put more emphasis on reducing racial biases. These different opinions can be reconciled in a transparent and principled manner under our framework. Thus, our framework should promote social discourses and increases people's trust in using AI in high-stake decision-making scenarios, which in turn increases efficiency and fairness in the society.

\bibliographystyle{ACM-Reference-Format}
\bibliography{citations}

\end{document}